# Non-invasive measuring method of skin temperature based on skin sensitivity index and deep learning

Xiaogang Cheng [1,4,5], Bin Yang [2,3], Kaige Tan [4], Erik Isaksson [4], Liren Li [6], Anders Hedman [4], Thomas Olofsson [3] and Haibo Li [1,4]

[1] College of Telecommunications and Information Engineering, Nanjing University of Posts and Telecommunications, Nanjing, 210003, China
[2] School of Building Services Science and Engineering, Xi'an University of Architecture and Technology, Xi'an, 710055, China
[3] Department of Applied Physics and Electronics, Umeå University, 90187 Umeå, Sweden
[4] Royal Institute of Technology (KTH), Stockholm, 10044, Sweden
[5] Computer Vision Laboratory (CVL), Swiss Federal Institute of Technology (ETH), Zürich, 8092, Switzerland
[6] School of computer science and technology, Nanjing Tech University, Nanjing, 211816, China

**Featured Application:** The NISDL method proposed in this paper can be used for real time non-invasively measuring human skin temperature, which reflect human body thermal comfort status and can be used for control HVAC devices.

**Abstract:** In human-centered intelligent building, real-time measurements of human thermal comfort play critical roles and supply feedback control signals for building heating, ventilation, and air conditioning (HVAC) systems. Due to the challenges of intra- and inter-individual differences and skin subtleness variations, there is no satisfactory solution for thermal comfort measurements until now. In this paper, a non-invasive measuring method based on skin sensitivity index and deep learning (NISDL) was proposed to measure real-time skin temperature. A new evaluating index, named skin sensitivity index (SSI), was defined to overcome individual differences and skin subtleness variations. To illustrate the effectiveness of SSI proposed, two multi-layers deep learning framework (NISDL method I and II) was designed and the DenseNet201 was used for extracting features from skin images. The partly personal saturation temperature (NIPST) algorithm was use for algorithm comparisons. Another deep learning algorithm without SSI (DL) was also generated for algorithm comparisons. Finally, a total of 1.44 million image data was used for algorithm validation. The results show that 55.6180% and 52.2472% error values (NISDL method I, II) are scattered at [0 °C, 0.25 °C), and the same error intervals distribution of NIPST is 35.3933%.

## 1. Introduction

The energy consumption of urban residential and commercial housing accounts for 21% of the global energy consumption, and half of the housing consumption is generated by heating, ventilation and air conditioning (HVAC) systems [1,2]. Furthermore, one of the most important reasons of energy waste is that the actual thermal requirements of indoor occupants are ignored, with the result that overheating and overcooling occur often. Fortunately, real-time thermal comfort perception can provide useful signals to HVAC systems for achieving energy saving and human-centered intelligent control. Therefore, many researchers have been studying thermal comfort measurements for indoor environments in the past decades. Many methods were generated, including the questionnaire survey method [3–5], environmental measurement method [6,7] and invasive measuring method of human body physiological parameters [8–18]. In recent years, semi-invasive measuring method [19,20] and non-invasive measuring method [21–23] for human body physiological parameters were also generated. For example, in [19,20], an infrared sensor was fixed on the frame of eyeglasses in order to measure skin temperature. In [21], a normal vision sensor was also used for measuring skin temperature and two non-linear models were trained. In [22,23], Kinect was used for recognizing human poses or indoor locations, and then human thermal comfort and dynamic metabolism were estimated, respectively. All these methods are meaningful attempts. However, due to the challenges of measuring thermal comfort which are (1) skin subtleness variation [24], (2) inter-individual differences [14,25] and (3) temporal intra-individual differences [14,26], there is still no satisfactory method for perceiving human thermal comfort.

To overcome the aforementioned challenges, skin sensitivity index (SSI) was defined in this paper. The SSI is strongly related to skin temperature. A Non-Invasive measuring method of skin temperature based on SSI and Deep Learning was proposed, hereinafter referred to as NISDL. Two different deep learning methods of NISDL have been designed and trained, respectively, which are NISDL method I and II. The main difference between them is that the location of SSI participation in the neural network training is not the same. A total of 1.44 million images were collected for 16 Asian female subjects, and this 'big data' was used for algorithm validation.

The main contributions of this paper are:

(1) Skin sensitivity index (SSI) was proposed for describing individual sensitivity of thermal comfort, and the index was combined with skin images for deep learning network training.

(2) A novel non-invasive measuring algorithm (NISDL) based on SSI was proposed, with two different frameworks of NISDL having been designed for real-time thermal comfort measurement.

(3) A deep learning algorithm without SSI was also generated and trained. Two comparisons were made: 1) comparison between data-driven methods (deep learning) and model-driven methods (linear models); 2) comparison of measuring effects in the case of SSI participation in training and non-participation in training.

The rest of this paper is organized as follows. Section 2 introduces the related work about thermal comfort. In section 3, the research methods, including SSI computation, subjective experiments and NISDL methods, are introduced. The results and discussion are shown in sections 4 and 5. Finally, section 6 gives the conclusion.

**2. Related work**

Since the 1970s, Fanger has explored human thermal comfort and conducted many kinds of subjective experiments. Based on this, he eventually established what is known as Fanger's theory [27]. From then on, many studies about thermal comfort were carried out.

Questionnaire surveys are good as a method to understand the inner feeling of an occupant. With the development of the internet, online questionnaire surveys can also be generated [3,4]. However, it is inconvenient and also difficult to guarantee that occupants will continue to give feedback on their personal thermal feelings [5]. Therefore, the environment measurement method was also adopted in building industry [6]. In this kind of method, some objective parameters, such as indoor temperature, airflow and humidity, are often measured. Unfortunately, the goal of the environment measurement method is to meet the thermal comfort needs of a majority of indoor occupants. Therefore, the thermal feelings of a minority were ignored. For overcoming this drawback, a kind of nonlinear autoregressive network, still belonging to the environment measurement method, was generated to predict indoor temperature [7]. In fact, human thermal comfort is complicated, and with constant indoor parameters it is difficult to meet each individual's requirement of thermal comfort. As such, some researchers study physiological measurement methods, including the invasive measuring method, semi-invasive measuring method and non-invasive measuring method.

For the invasive measuring method, skin temperature and heart rate are usually the measured parameters. Wang and Nakayama [8,9] made early attempts at skin temperature around the human body. Liu [10] proposed a method to estimate mean values of skin temperature. A total of 22 subjects were invited for subjective experiments. The data, being local skin temperatures and electrocardiograms, were collected. Takada [11] presented a multiple regression equation to predict skin temperature in non-steady state. The multiple regression equation was considered as a function of mean skin temperature. Wrist skin temperatures and upper extremity skin temperatures were also adopted to estimate human thermal sensation, respectively [12,13]. Chaudhuri [14] presented a predicted thermal state (PTS) model, and capture of peripheral skin temperature. Further, body surface area and clothing insulation were used for analyzing inter- and intra-individual differences. As to the thermal comfort study using heart rate, based on physiological experimentation, Yao [15] investigated the relationship between heart rate variation (HRV) and electroencephalograph (EGG). Moreover, Dai, Chaudhuri and Kim [16–18] combined machine learning with invasive measurement. All of [16-18] used an SVM classifier to predict human thermal sensation in different experiments.

For the semi-invasive measuring method, Ghahramani [19] used an infrared sensor to estimate skin temperature of different face points. The infrared sensor was mounted on the frame of eyeglasses. Based on this, a hidden Markov model was constructed to capture personal thermal sensation [20].

In practical application, invasive and semi-invasive measurement are both difficult to apply widely. The reason is that an occupant needs to wear a sensor which is uncomfortable and it is also not in line with the goal of human-oriented intelligent buildings. For this reason, a kind of non-invasive measuring method was studied in [21]. Based on vision sensor, Cheng [21] extracted the saturation (S) channel from skin images and constructed two saturation-temperature models to estimate skin temperature. The two models are the non-invasive measuring method of thermal comfort based on saturation-temperature (NIST) and the non-invasive measuring method of thermal comfort based on partly saturation-temperature (NIPST). Alan [22] proposed a non-invasive measuring method based on human poses. A total of 12 poses of thermal comfort were defined and Kinect was adopted to estimate human skeleton and poses. Further, Dziedzic [23] also used Kinect to predict human thermal sensation and dynamic metabolic rate.

In the past decades, the measurement of thermal comfort has been mainly focused on traditional experimental science. Taleghani, de Dear, Rupp and Djamila [28–31] also reviewed the study of thermal comfort. Some databases (such as ASHRAE RP-884) and international standards (such as ASHRAE 55-2013 standard, European EN15251 standard) were discussed. With the development of machine learning (ML) and computer vision (CV), some thermal comfort perception methods based on ML and CV were proposed. Support Vector Machine (SVM) are often used for analyzing existing databases (RP-884) and captured environmental parameters [24,32–33]. Further, Peng [26] use unsupervised and supervised learning to predict occupants' behavior, applied to three types of offices which are single person offices, multi-person offices, and meeting rooms. Li [34] proposed a fuzzy model to predict thermal sensation, skin temperature and heart rate considered as objective parameters. For avoiding overheating, Cosma [35] extracted data from multiple local body parts and analyzed them with four kinds of machine learning algorithms, including SVM, Gaussian process classifier (GPC), k-neighbors classifier (KNC) and random forest classifier (RFC).

The kinds of machine learning adopted in [24,26,32–35] are traditional algorithms. In recent years, use of deep neural networks is boosting [36,37]. In addition, a kind of subtleness magnification technology was presented [38,39]. These provide new directions and opportunities for the measurement of human thermal comfort. Based on it, a novel non-invasive measuring method was generated which will be introduced as follows.

## 3. Research methods

*3.1. Subjective physiological experiments*

**Subjects data and chamber environments.** 16 human subjects were invited for experiments and the resulting data volume is 1.44 million images. The experiments were conducted in a chamber with controllable indoor air temperature and relative humidity. The corresponding dry-bulb air temperature is 22.2 ± 0.2 °C and the relative humidity is 36.9 ± 2.5%. The resolution of vision sensor used for capturing video is 1280 × 720. The iButton, model DS192H with uncertainty ± 0.125 °C, was used for measuring skin temperature from the back of subject's hand. All the subjects are Asian females with average age of 23.9 ± 3.9 years, average weight of 52.2 ± 6.5 kg, and body mass index (BMI) 19.9 ± 2.2 kg/m$^2$.

**Experimental procedures.** There are mainly three steps in subjective physiological experiments. (1) Preparation stage: The indoor environment parameters were measured and controlled to a suitable level. When the subjects came into the chamber, they should have a rest for 10 minutes for adaptation. At the same time, warm water with constant temperature (45 °C) was prepared. (2) Thermal stimulus: After 10 minutes' adaptation, subjects were asked to immerse hands into the water with 45 °C. The whole thermal stimulus process lasted for 10 minutes. (3) Big data collection: After 10 minutes of stimulus, subjects were asked to sit next to the data collection desk and put her pairs of hands under the vision sensor. The back of the hand is faced up and the data is collected for 50 minutes. At the same time, skin temperature sensor (iButton) was attached to the back of one hand. The corresponding sampling interval is 1 minute. It should be noted that, based on piecewise stationary time series analysis [40], linear interpolation was adopted in this paper, and 11 points were interpolated into 1 minute for real skin temperature captured by iButton.

*3.2. Skin sensitivity index*

**SSI definition.** When human body encounters thermal stimuli, blood circulation will change which will also be reflected in skin's color and texture. In [21], based on the HSV (hue, saturation, value) color space, the S channel was extracted and a linear ST model was established.

$$T = k_i \times S + b_i \quad (1)$$

Where k reacts to the change rate of skin temperature. In this paper, k is defined as the skin sensitivity index (SSI). SSI is a high weight coefficient in skin temperature changes and SSI reflects the skin sensitivity level to external thermal stimuli.

**SSI computing.** Based on subjective physiological experiments, real skin temperature can be obtained by iButton. The images were also collected from subjects' hand. Therefore, the SSI can be calculated. The steps are as follows: (1) Extracting each frame from captured video; (2) Segment region of interest (ROI); (3) Extracting S channel from ROI images and computing mean values of S for each ROI image; (4) Search SSI value based on real skin temperature and S for each subject.

*3.3. NISDL algorithm*

In this paper, considering that SSI is high weight coefficient for non-invasive thermal comfort measurement, it will improve the prediction accuracy of skin temperature. Based on SSI, the NISDL algorithm was introduced in this paper. Furthermore, to validate the effectiveness of SSI, two kinds of deep learning frameworks (NISDL method I, II) have been constructed. The main difference between NISDL method I and II is that the location where the network invokes SSI to participate in the model training is different. The NISDL algorithm constructed is introduced as follows.

**Video pre-processing.** In fact, skin texture variation is subtle which is difficult to be perceived. In this paper, for magnifying this kind of subtleness variation, an image subtleness magnification technology is adopted which is Euler Video Magnification (EVM) [38,39]. Based on EVM, let c (x, t) denote skin images which is subtly varied with time t. Supposed that the variation function is formula (2) [38,39].

$$C(x, t) = F(x + h(t)) \quad (2)$$

Where, h (t) is variation degree, F is a function which constructs the relationship between C (x, t) and h (t). If the skin image C (x, t) is magnified, and first-order Taylor series expansion can be handed to F.

$$C(x, t) = F(x + (1+ \xi) * h(t)) \qquad (3)$$

Where ξ is the magnification coefficient which can be set based on practical application. According to formula (3), only the variation part was magnified to a magnitude of 1+ξ, while the other part of skin texture is not magnified. Therefore, the invisible texture variation is made to be visible.

It should be noted that de-noise processing should be handled before EMV processing. Further, after the video is magnified, ROI is selected and cropped from each frame. The ROI images are imported into NISDL method I and II for model training.

**NISDL method I.** As shown in Figure 1, SSI values are used as input data and imported into the deep learning network at the very beginning. The ROI images were also combined with SSI values in the first step. According to the size of ROI images, the SSI value of each ROI image was expanded into a matrix. The matrix is considered as a channel and combined with the 3 channels of ROI images.

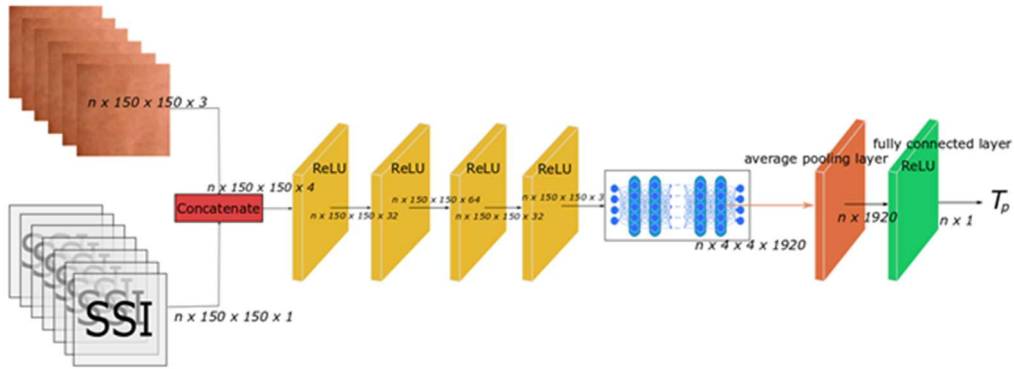

**Figure 1.** NISDL method I. (SSI values were combined with skin images firstly.)

The merged data between ROI images and SSI values above is inputted into four convolution layers which are used for dimensionality reduction. In NISDL method I, the DenseNet201 is adopted for features extraction. The last two layers of DenseNet201 are not suitable for skin temperature measurement, hence the two layers are removed. The reason is that the activation function of the last layers is softmax. Instead of these two layers, an average pooling layer and a fully connected layer are added behind denseNet201. Based on the deep learning networks designed above, n ROI images and n SSI values were inputted into NISDL method I. Therefore, n skin temperatures can be obtained.

**NISDL method II.** Figure 2 is the deep learning framework of NISDL method II. In this method, the ROI images and SSI values were processed for features extraction. Subsequently, the two kinds of features were combined in the second half of the whole framework. An average pooling layer and the DenseNet201 (excluding the last two layers) were also used for features extraction of ROI images. For SSI values, a convolution layer and an average pooling layer were adopted for feature extraction and dimensionality reduction. After features combination, three fully connected layers were constructed in NISDL method II. Therefore, the skin temperature can be obtained. The algorithm in details, including NISDL method I and II, are shown in Table 1.

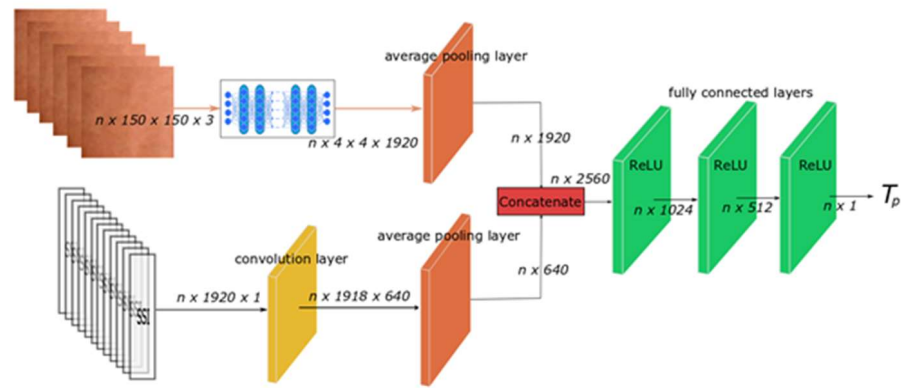

**Figure 2.** NISDL method II. (SSI values and skin images were processed by deep learning networks. Two kinds of features were extracted and the features were combined.)

**Evaluation metric.** For assessing NISDL algorithm constructed in this paper, the absolute error is adopted.

$$Error = |T_p(i) - T_r(i)| \quad i = 1, 2, 3, \ldots \qquad (5)$$

Where $T_p(i)$ is the prediction values of skin temperature and obtained from the proposed NISDL algorithm. $T_r(i)$ is the real value of skin temperature and captured by iButton. The parameter $i$ denotes the particular ROI image.

**Algorithms for comparison.** Two algorithms are used for comparison in this paper: (1) DL algorithm. The commonality between NISDL method I and II is that they all use SSI for model training. For validating the effectiveness of NISDL algorithm (with SSI), we remove the SSI and corresponding hidden layers for SSI features extraction from NISDL method II, so that it will be another deep learning network (without SSI) and named DL algorithm hereinafter. (2) NIPST algorithm. DL algorithm, NISDL method I and II are all nonlinear methods and data driven methods with deep learning networks. For further validating NISDL method I and II, the NIPST algorithm is also used for algorithm comparison which is a linear and model driven method.

Table 1. Non-invasive measuring method of skin temperature based on skin sensitivity index.

| Algorithm: the NISDL algorithm |
| --- |
| **Output:** NISDL model (*.*h5*), skin temperature (°C) |
| **Step:** |
| 1. Video pre-processing |
|   (1) De-noise and handle subtleness magnification for captured video. |
|   (2) Magnification coefficient *ξ* is 10 (formula (3)). |
|   (3) Extracting region of interest (ROI) from each frame of video, the size is 150×150. |
| 2. Making label |
|   (1) Making numerical interpolation for skin temperatures captured by iButton. |
|   (2) Uniform interpolation is adopted, plus 11 points / min. |
|   (3) Establish a correspondence table between ROI images and skin temperatures (after interpolation). |
| 3. Algorithm training |
|   (1) Commonality between NISDL method I and II |
|     1) Training set and test set ratio: 12 : 4. |
|     2) Validation set: 500 images. |
|     3) During network training, 32 images/batch, epoch is 8. |
|     4) Training ~30000 images, validate once. |
|     5) Activation function: ReLU |
|   (2) NISDL method I |
|     1) SSI values were concatenated with ROI images in the first steps. |
|     2) Convolutional kernel: 1×1. |
|   (3) NISDL method II |
|     1) Features are extracted from SSI values and ROI images, respectively. |
|     2) Concatenating the two kind of features in the second half of network. |
|     3) Convolutional kernel: 3×1. |
| 4. Optimizing model parameters |

## 4. Results

16 subjects were invited for subjective physiological experiments and a total of 1.44 million images were captured. Based on this, the NISDL algorithm was validated and compared with the NIPST algorithm and the DL algorithm.

**Hardware parameters.** For this paper, a computer with a GPU was used for images processing and algorithm validation. The GPU is GeForce GTX TITAN X, the CPU is Intel core i5-4460 CPU@3.2Ghz X 4, the RAM is 16G and the word size is 64bit.

**Training of NISDL method I.** The size of ROI images are n×150×150×3, and the size of expanded SSI vlues are n×150×150×1. The SSI matrix was considered as a channel and concatenated with ROI images, so that the result is n×150×150×4. The activation function of four convolution layers, shown in Figure 1, are Rectified Linear Units (ReLU) and the size of convolution kernel is 1×1. DenseNet201 was used for feature extraction and its output is a matrix with size of n×4×4×1920. Based on this, two hidden layers are constructed and the size of the last layer, being a fully connected layer, is 1920×n.

**Training of NISDL method I.** The size of the expanded SSI matrix is n×1920×1, which differs from that of DISDL method I. The corresponding convolution kernel is 3×1. The features of SSI were extracted by two hidden layers, and the features of ROI images were extracted by DenseNet201. As shown in Figure 2, in the second half of framework, the two kinds of features are concatenated. In order to ensure that the SSI features have a suitable influence on network training (moderate, not too big or too small), the size of ROI image features is set as n × 1920, and size of SSI features is set as n × 640 (triple relationship). Finally, the size of last three hidden layers are 2560×1024, 1024×512 and 512×1, respectively.

**Commonality between NISDL method I and II.** During network training, the same parameters of NISDL method I and II are shown as follows. Based on data of 1.44 million images, the ratio of training set and test set is 12:4, the number of validation set is 500. The epoch is 8 which means that the training set was trained 8 times. The input data batch is 32. When the error of validation set is less than 0.46 °C, the corresponding model (*.h5) will be saved. Further, when 30, 0000 images of training set were trained, the corresponding model (*.h5) will also be saved. After the generation of the, the test set images were inputted into generated model, so that the prediction values of skin temperatures could be obtained.

**Quantitative comparison.** The prediction values of skin temperature are shown in Figure 3. The set of values obtained from iButton is ground truth. The corresponding error statistics, including mean, median, are shown in Fig. 4 which is a box-whisker plot. The mean values of NISPT, DL, NISDL method I and II in °C are 0.5793, 0.3594, 0.3351 and 0.2647, respectively. In addition, the median values of them in °C are 0.3430, 0.3085, 0.2381 and 0.2282, respectively. It was shown that deep learning methods (DL, NISDL method I and II) are all better than the nonlinear model (NIPST) and further that the method with SSI (NISDL method I and II ) is better than the method without SSI (DL).

In this paper, the error distributions are given in Figure 5 and Table 2. The errors of DL, NISDL method I and II are mainly concentrated in the range of 0 °C and 0.75 °C. NISDL is better than DL, because two error percentages of NISDL corresponding to [0, 0.25) are 52.2472% and 55.6180%. In addition, the error percentages of NISDL corresponding to [0.25, 0.5) and [0.5, 0.75) are less than that of DL. The error percentage of NIPST is increased from the interval of [0.75, 0.1, meaning that the performance of NIPST is worse than DL and also NISDL methods I and II.

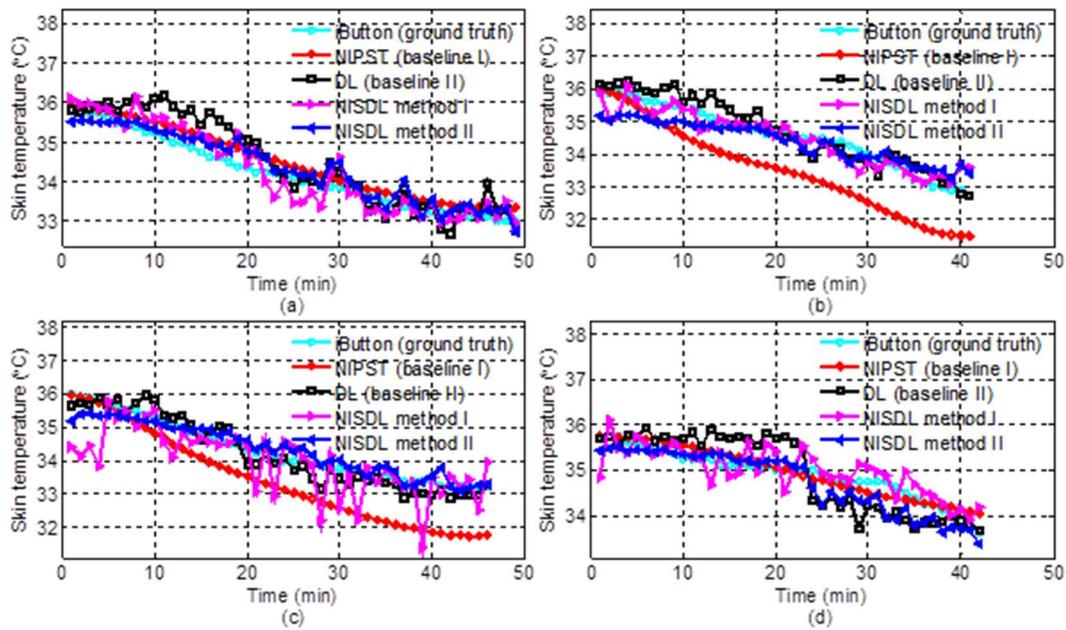

**Figure 3.** Skin temperatures comparison between ground truth, baseline and NISDL (Images of 12 subjects (0.96 million images) were training set; images of 4 subjects (0.32 million images) were test set).

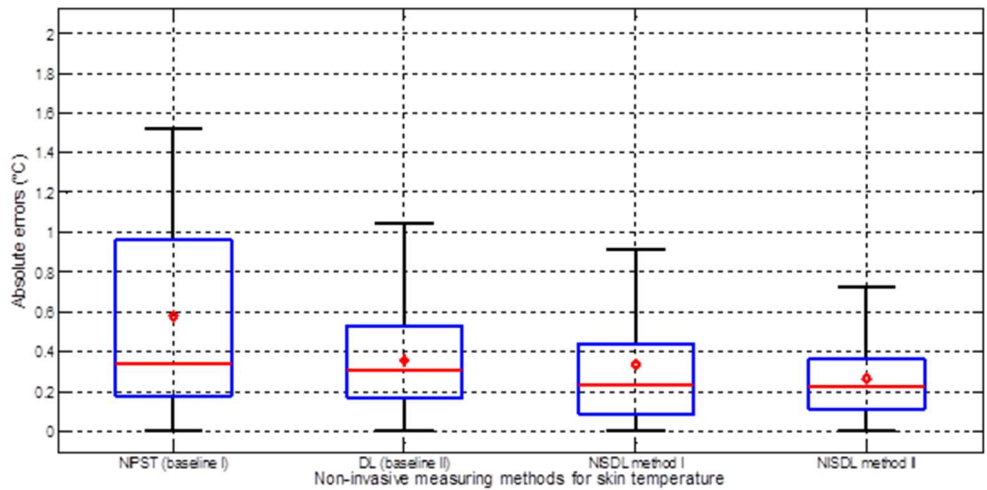

**Figure 4.** Error statistics (box-whisker plot) comparison between baseline and NISDL (a. NIPST was published in [21]. b. Skin images were trained directly by DenseNet201 to obtain a model and predict skin temperature, and SSI was not involved. c. NISDL method I and NISDL method II all belong to NISDL. The main difference is that SSI values and skin images are combined at different times.)

**Table 2.** Absolute error distribution.

| Absolute error (°C) | NIPST (baseline, %) | DL (baseline, %) | NISDL Method I (Fig. 1, %) | NISDL Method II (Fig. 2, %) |
| --- | --- | --- | --- | --- |
| [0, 0.25) | 35.3933 | 37.6404 | 52.2472 | 55.6180 |
| [0.25, 0.5) | 24.1573 | 35.9551 | 28.0899 | 30.3371 |
| [0.5, 0.75) | 5.0562 | 19.1011 | 10.1124 | 11.7978 |
| [0.75, 1.0) | 11.2360 | 5.6180 | 3.3708 | 2.2472 |
| [ >1.0) | 24.1573 | 1.6854 | 6.1798 | 0 |

## 5. Discussion

**Situation of overcoming challenges.** NISDL has overcome the three challenges, mentioned in section 1, to some extent. A kind of subtleness magnification technology, which is Euler Video Magnification (EVM), was used for magnifying the skin texture variation, so that the challenges '(1)' given in section 1 can be overcome. For overcoming challenges '(2)' which is inter-individual difference, skin sensitivity index (SSI) was proposed and SSI is related with skin saturation. Fig. 4-5 shows that the performance of NISDL algorithm with SSI is better than that of DL without SSI. In practical application, skin images will be captured in real-time (30 frames/s), the skin temperature variation can always be obtained. Therefore, challenges '(3)' proposed in Section 1 can be overcome. Furthermore, piecewise stationary time series analysis was adopted in this paper for overcoming challenges '(3)'. Considering operability, the breakpoint interval of piecewise stationary signal is set to 5 seconds, supposing, e.g., that the skin temperature is constant value in 5 seconds.

**The deep learning framework.** In this paper, DISDL method I and II are all belonging to the deep learning method. In addition, DL generated for algorithm comparison is also a deep learning method. Figure 4 and Figure 5 show that the deep learning method is better than a linear model (NIPST). From the perspective of deep learning, the main reason is that big data is adopted.

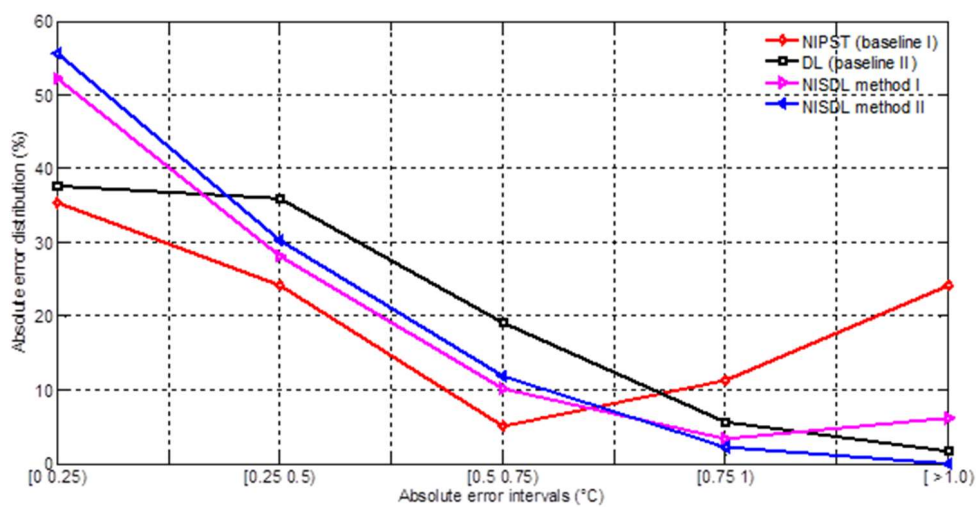

**Figure 5.** Error distribution comparison between baseline and NISDL (a. NIPST was published in [21]. b. Skin images were used in training directly by DenseNet201 to obtain a model and predict skin temperature, and SSI was not involved. c. NISDL method I and NISDL method II all belong to NISDL. The main difference is that SSI values and skin images are combined at different times.).

**SSI proposed.** Although NISDL and DL are all better than NIPST, there are still big gap between NISDL and DL. Figure 4, Figure 5 and Table 2 also show that NISDL is better than DL. The main reason is that SSI is used in NISDL. The NISDL method II is also shown to be better than NISDL method I. Further, when SSI features are extracted and concatenated with ROI images features in the second half of network, the performance will be better.

**Reasons of designing two frameworks for NISDL.** Some researcher may ask, why do we design NISDL methods I and II together? The main reason is that we want to extensively confirm the effectiveness of SSI. In this paper, SSI participates in network training from different locations, and the results are all good. When the SSI values are removed from network (DL), the corresponding performance decreased significantly. Based on this, we can know that SSI is helpful for predicting skin temperature through deep learning networks.

**Practical Application.** Some researcher may argue that the method proposed in this paper still can not be applied in practice right now. In fact, a method is always being gradually improved. For example, the NISDL proposed in this paper is better than NIPST which was proposed in 2017. In addition, when more diverse data is captured and used for model training, the performance of NISDL will be better.

Some researcher may argue that the infrared sensor also can be used for measuring skin temperature, so why do we use a vision-based method? In fact, the study [19,20] focused on thermal comfort measurement with an infrared sensor. However, the measurement accuracy is limited. Beyond this, there are other drawbacks in the infrared based measuring method: 1) Distance. The infrared sensor should be placed close to occupant. 2) Cost. The infrared sensor with high accuracy is expensive and the accuracy of an infrared sensor with low cost is also low. 3) Information is limited. The infrared sensor is similar with the 'human tactile organ' which only can get limited thermal comfort information. But the vision-based methods are similar to the 'human eye'. Some information, such as poses of thermal comfort, can be obtained and analyzed by a vision-based method, but not by an infrared-based method. Based on these three drawbacks, the infrared-based method is challenging to be widely applied in practice.

Furthermore, some other researchers may say that vison-based non-invasive measuring method may have concerns related to personal privacy. In fact, there are at least two options to protect personal privacy issues: 1) Switch button. Based on this switch button, any customer can choose to accept or reject the implementation of real-time personal service of thermal comfort. 2) Information selection. In future practical application, only the information about human thermal comfort be processed and saved, while other information will be discarded in real-time. Therefore, from the perspective of human-centeredness, the NISDL algorithm proposed in this paper is helpful.

## 6. Conclusions

In this paper, a kind of non-invasive measuring method based on skin sensitivity index for thermal comfort (NISDL) is proposed. For validating the effectiveness of SSI, two different deep learning frameworks with SSI were designed. A total of 1.44 million images were used for algorithm validation. The conclusions can be summarized as follows.

(1) SSI is a good and high weight parameter in non-invasive measurement of skin temperature based on deep learning network.

(2) The location of SSI participation in NISDL network training has little of impact on measuring performance of skin temperature. Of course, if the SSI features are extracted firstly, and then merged with the features of ROI images, the corresponding effect is slightly better.

(3) The NISDL method proposed in this paper can be used for measuring thermal comfort and more diverse data can help it to improve the measuring accuracy.

In practical application, the inter-difference is very big. How to define and calculate suitable SSI will affect the measuring results. Further, more diverse data comparison is required to improve the algorithm robustness. These will be our research direction in the near future.

analysis, Xiaogang Cheng and Bin Yang; writing—original draft preparation, Xiaogang Cheng; writing—review and editing, Bin Yang and Anders Hedman.

analysis, Xiaogang Cheng and Bin Yang; writing—original draft preparation, Xiaogang Cheng; writing—review and editing, Bin Yang and Anders Hedman.

**Funding:** This research was funded by the National Natural Science Foundation of China (No. 61401236), the Jiangsu Postdoctoral Science Foundation (No. 1601039B), the Key Research Project of Jiangsu Science and Technology Department (No. BE2016001-3).

**Acknowledgments:** The authors thank William T. Freeman (MIT) for providing his MATLAB code of Euler Video Magnification (EVM).

**Conflicts of Interest:** The authors declare no conflict of interest.